\documentclass[letterpaper, 10 pt, conference]{ieeeconf}  

\IEEEoverridecommandlockouts                              

\usepackage{times}
\usepackage{epsfig}
\usepackage{graphicx}
\usepackage{caption}
\usepackage{amsmath}
\usepackage{amssymb}
\usepackage{array,multirow,booktabs}
\usepackage{hyperref}
\usepackage{capt-of,etoolbox}
\usepackage{amsfonts}
\usepackage{hyperref}

\graphicspath{{./figures/}}
\usepackage{color}
\usepackage[table]{xcolor}
\definecolor{LightCyan}{rgb}{0.88,1,1}
\definecolor{Green}{HTML}{99FF66}

\title{Towards real-time unsupervised monocular depth estimation on CPU}

\author{Matteo Poggi$^{1}$, Filippo Aleotti$^{2}$, Fabio Tosi$^{1}$, Stefano Mattoccia$^{1}$
\thanks{$^{1,2}$Department of Computer Science and Engineering (DISI), University of Bologna, 40136 Bologna, Italy. $^{1}$\{m.poggi, fabio.tosi5, stefano.mattoccia\}@unibo.it, $^{2}$filippo.aleotti@studio.unibo.it}}
\begin{document}

\maketitle
\thispagestyle{empty}
\pagestyle{empty}

\begin{abstract}
Unsupervised depth estimation from a single image is a very attractive technique with several implications in robotic, autonomous navigation, augmented reality and so on. This topic represents a very challenging task and the advent of deep learning enabled to tackle this problem with excellent results. However, these architectures are extremely deep and complex. Thus, real-time performance can be achieved only by leveraging power-hungry GPUs that do not allow to infer depth maps in application fields characterized by low-power constraints. To tackle this issue, in this paper we propose a novel architecture capable to quickly infer an accurate depth map on a CPU, even of an embedded system, using a pyramid of features extracted from a single input image. Similarly to state-of-the-art, we train our network in an unsupervised manner casting depth estimation as an image reconstruction problem.
Extensive experimental results on the KITTI dataset show that compared to the top performing approach our network has similar accuracy but a much lower complexity (about 6\% of parameters) enabling to infer a depth map for a KITTI image in about 1.7 s on the Raspberry Pi 3 and at more than 8 Hz on a standard CPU. Moreover, by trading accuracy for efficiency, our network allows to infer maps at about 2 Hz and 40 Hz respectively, still being more accurate than most state-of-the-art slower methods. To the best of our knowledge, it is the first method enabling such performance on CPUs paving the way for effective deployment of unsupervised monocular depth estimation even on embedded systems.


\end{abstract}


\section{Introduction}

Several application fields such as robotic, autonomous navigation, augmented reality and many others can take advantage of accurate and real-time depth measurements. Popular \emph{active} sensors such as LIDAR, Kinect, Time-of-Flight (ToF) infer depth by perturbing the sensed environment according to different technologies. Despite their effectiveness in specific circumstances (e.g., the Kinect for close range indoor deployment), \emph{passive sensors} based on binocular/multi-view stereo, structure from motion and, more recent, monocular depth sensors are very attractive. In fact, they are potentially cheaper, smaller and more lightweight than active sensors. Moreover, passive depth sensors don't have moving parts like LIDAR and don't require to perturb the sensed environment thus avoiding interference with other devices.


\begin{figure}[t]
\begin{tabular}{c}
\includegraphics[width=7.8cm]{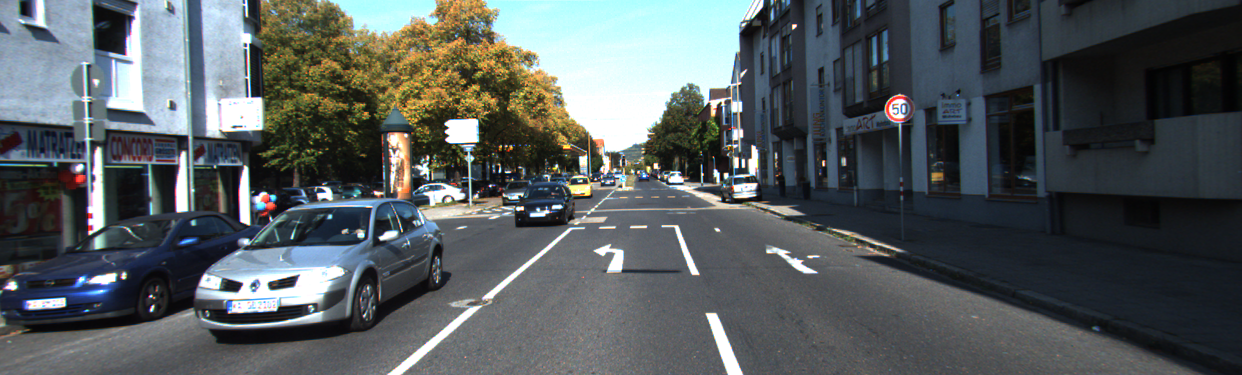} \\
\includegraphics[width=7.8cm]{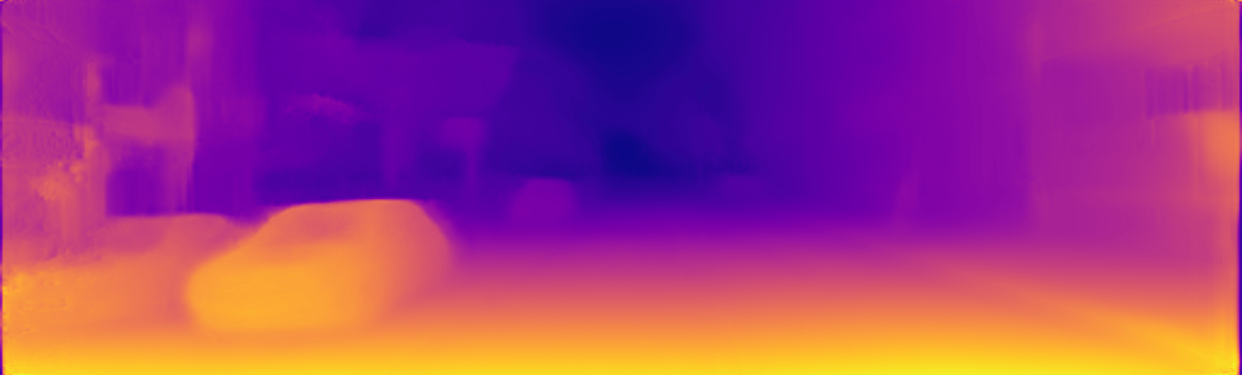} \\
\includegraphics[width=7.8cm]{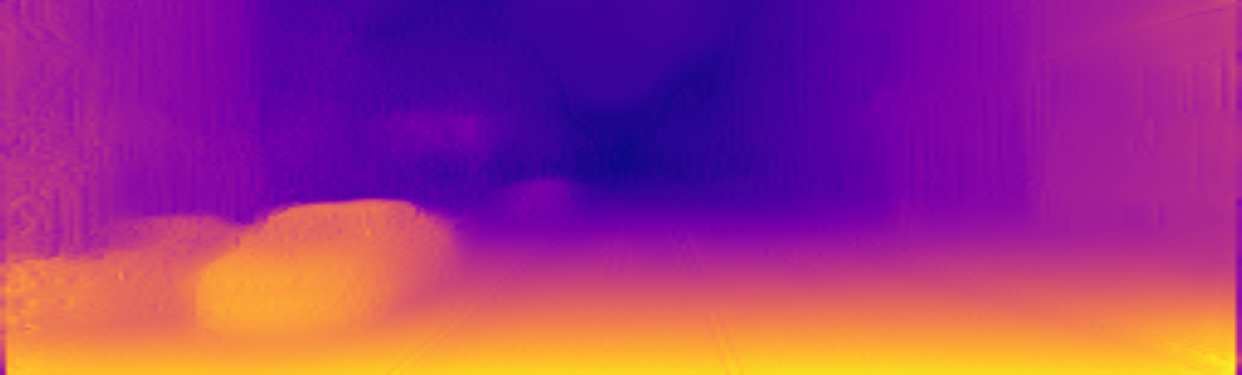} \\
\end{tabular}
\caption{(Top) Input image from KITTI dataset \cite{geiger2012we}. Qualitative comparison between state-of-the-art unsupervised monocular depth estimation method \cite{godard2017unsupervised} (Middle) and the proposed PyD-Net architecture (Bottom). Our model runs in real-time on standard CPUs and takes, in its most accurate configuration reported in this figure, 1.7 s on the low-power ARM CPU of the Raspberry Pi 3 with an overall power consumption, including a web-camera, of about 3.5 W.}
\end{figure}

The literature concerned with passive depth sensors is large but in recent years most methods have been outperformed by approaches leveraging on Convolutional Neural Networks (CNNs). In particular, CNNs allowed to effectively increase the accuracy of passive techniques by casting the depth data generation as a supervised learning task. CNNs also enabled depth estimation from a single input image thus avoiding acquisitions from multiple view points for this purpose. While some seminal works concerned with monocular depth estimation \cite{ladicky2014pulling,eigen2014depth,liu2016learning} require a large amount of training samples with depth labels, more recent works \cite{zhou2017unsupervised,godard2017unsupervised} exploit unsupervised signals in form of image reconstruction losses to train CNNs on monocular sequences \cite{zhou2017unsupervised} or stereo pairs \cite{godard2017unsupervised} required only for training and not for inference. With this latter strategy, difficult to source ground-truth labels are replaced with standard imagery enabling to collect training samples easily and in large amounts. Nevertheless, current architectures for monocular depth estimation are very \emph{deep} and complex; for these reasons they require dedicated hardware such as high-end and power-hungry GPUs. This fact precludes to infer depth from a single image in many interesting applications fields characterized by low-power constraints (e.g. UAVs, wearable devices, ...) and thus in this paper we propose a novel architecture for accurate and unsupervised monocular depth estimation aimed at overcoming this issue. By building our deep network inspired by the success of pyramidal architectures in other fields \cite{he2014spatial,ranjan2017optical} we are able to decimate the amount of parameters w.r.t. state-of-the-art solutions thus dramatically reducing both memory footprint and runtime required to infer depth. We call our model \underline{Py}ramidal \underline{D}epth \underline{Net}work (PyD-Net) and we train it in unsupervised manner as proposed in \cite{godard2017unsupervised}, representing the top-performing method in this field. Compared to such work, our model is about 94\% smaller enabling on CPUs a notable speed-up at the cost of a slightly reduced depth accuracy. Moreover, our proposal outperforms other state-of-the-art methods. 
Our design strategy enables the deployment of PyD-Net even on embedded devices, such as the Raspberry Pi 3, thus allowing to infer a full depth map at about 2 Hz using less than 150 MB out of 1 GB memory available in such inexpensive device. To the best of our knowledge, our proposal is the first approach enabling fast and accurate unsupervised monocular depth estimation on standard and embedded CPUs.

\section{Related Work}

Although depth estimation from images has a long history in computer vision \cite{scharstein2002taxonomy}, methods using a single image are much more recent and mostly based on machine learning techniques. These works and other efficient end-to-end approaches for dense prediction tasks such as optical flow are relevant to our work.

\textbf{Supervised monocular depth estimation.} 
Saxena et al. \cite{saxena2009make3d} proposed Make3D, a patch-based model estimating 3D location and orientation of planes by means of a Markov Random Field framework. It suffers in presence of thin structures and does not process global context information because of its local nature. Liu et al. \cite{liu2016learning} trained a CNN to estimate depth from single camera, while Ladicky et al. \cite{ladicky2014pulling} included semantic information into their model to obtain more accurate predictions. In \cite{karsch2014depth} Karsch et al. obtained more consistent predictions by casting the problem as a nearest neighbor search with respect to depth images from a training set required at testing time. Eigen et al. \cite{eigen2014depth} deployed a multi-scale CNN trained on a large dataset to infer depth for each pixel in a single image. Differently from \cite{liu2016learning}, whose network was trained to compute more robust data terms and pairwise terms fed to a further optimization step, this approach directly infers the depth map.
Several works followed \cite{eigen2014depth} to improve its performance by means of CRF regularization \cite{li2015depth}, casting the problem as a classification task \cite{cao2017estimating}, designing more robust loss functions \cite{laina2016deeper} or using scene priors for joint plane normals estimation \cite{wang2015designing}.
Ummenhofer et al. \cite{ummenhofer2017demon} proposed DeMoN, a deep model to infer both depth and ego-motion from a pair of subsequent frames acquired by a single camera. 
Common to all these works is the supervised paradigm adopted for training, requiring a large amount of labeled data particularly crucial for successfully learn a robust depth representation from a single image.

\textbf{Unsupervised monocular estimation.} 
Other recent works exploit CNNs without using labeled data. In particular, Flynn et al. \cite{flynn2016deepstereo} proposed DeepStereo, a deep architecture trained on images acquired by multiple cameras to synthesize images from new view points. In the context of binocular stereo, given an input reference image, Deep3D by Xie et al. \cite{xie2016deep3d} generates the corresponding target view by learning a distribution over all possible disparities at each pixel on the source image. For training, an image reconstruction loss is minimized. Similarly, Garg et al. \cite{garg2016unsupervised} trained a network for monocular depth estimation using the same objective loss principle over a stereo pair, using Taylor approximation to make their loss linear and fully differentiable thus making their framework trainable in end-to-end manner but resulting in a more challenging objective function to optimize.

To overcome this issue, Godard et al. \cite{godard2017unsupervised} used a bilinear sampling \cite{jaderberg2015spatial} to generate images from depth predictions. At training time, the model learns to predict depth for both images of a stereo pair by processing reference image only, enabling a left-right consistency check when computing the loss signal to minimize and a simple post-processing step to obtain a more accurate prediction. Currently, this work represents state-of-the-art for monocular depth estimation. Poggi et al. \cite{3net18} improved \cite{godard2017unsupervised} with an interleaved training technique, simulating a trinocular setup out of a binocular stereo dataset allowing to obtain a more accurate model. 
While these methods require rectified stereo pairs for training, Zhou et al. \cite{zhou2017unsupervised} trained a model to infer depth from unconstrained video sequences by computing a reconstruction loss between subsequent frames and predicting, at the same time, the relative pose between them. This removes the requirement for stereo pairs, but produces a less accurate final model.

\textbf{Pyramidal networks for optical flow estimation.} Encoder-decoder architectures \cite{vincent2008extracting} have been widely adopted in computer vision when dealing with dense prediction. Most of them use skip connections between encoding and decoding parts to preserve fine details as done by U-net \cite{ronneberger2015u}. While these models count a large number of trainable parameters, pyramidal architectures recently proved to be very effective for optical flow \cite{ranjan2017optical,sun2017pwc}, outperforming U-net like architectures in terms of accuracy and, at the same time, decimating network parameters.

\section{Proposed method}

In this paper we propose a novel framework for accurate and unsupervised monocular depth estimation with very limited resource requirements enabling such task even on CPUs of low power devices. State-of-the-art architectures proposed for this purpose \cite{godard2017unsupervised} run in real time on high-end GPUs (e.g., Titan X), increasing the running time to nearly a second when running on standard CPUs and more than 10 s on embedded CPUs. Moreover, they count a huge number of parameters and thus require a large amount of memory at forward time.
For these reasons, real-time performance with such models are feasible only with high-end and power hungry GPUs.

To overcome this issue, we propose a compact CNN, enabling accuracy comparable to state-of-the-art, with very limited memory footprint at test time (i.e., $<$ 150 MB) and capable to infer depth at about 2 fps on embedded devices such as the Raspberry Pi 3 and tens of fps on standard CPUs whereas other methods are far behind.


To this aim some recent works in other fields have shown how classical computer vision principles, such as image pyramid, can be effectively adopted to design more compact networks. SpyNet \cite{ranjan2017optical} and PWC-Net \cite{sun2017pwc} are examples in the field of optical flow estimation with the latter representing state-of-the-art on MPI Sintel and KITTI flow benchmarks. The main difference with U-Net like networks is the presence of multiple small decoders working at different resolutions, directly on a pyramid of images \cite{ranjan2017optical} or features \cite{sun2017pwc} extracted by a very simple encoder compared to popular ones such as VGG \cite{simonyan2014very} or ResNet \cite{he2016deep}. Results at each resolution are up-sampled to the next level to refine flow estimation. This method allows for a large reduction in the number of parameters together with a faster computation in optical flow and we follow a similar strategy for our monocular depth estimation network depicted in Figure \ref{fig:architecture}.
To train PyD-Net we adopt the unsupervised protocol proposed by Godard et al. \cite{godard2017unsupervised} by casting depth prediction as an image reconstruction problem. For training unlabeled stereo pairs are required: for each sample, the left frame is processed through the network to obtain inverse depth maps (i.e., disparity maps) with respect to left and right images. These maps are used to warp the two input images towards each other and the reconstruction error is used as supervisory signal for back-propagation.

\section{PyD-Net architecture}

\begin{figure}
\centering
\includegraphics[width=6.2cm]{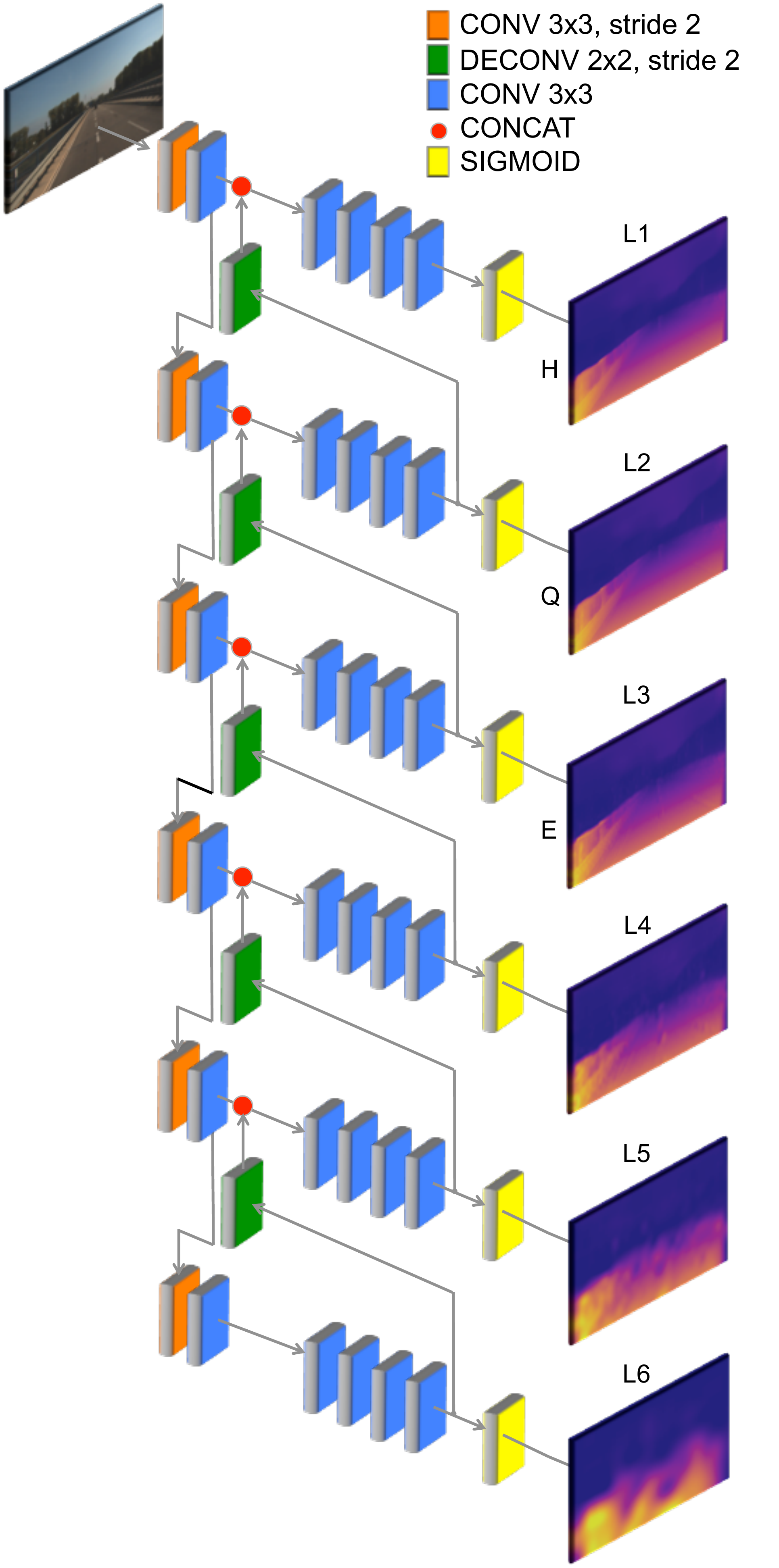}
\caption{PyD-Net architecture. A pyramid of features is extracted from the input image and at each level a shallow network infers depth at that resolution. Processed features are then up-sampled to the above level to refine estimation, up to the highest one.}
\label{fig:architecture}
\end{figure}

In this section we describe the proposed PyD-Net architecture depicted in Figure \ref{fig:architecture}, a network enabling results comparable to state-of-the-art methods but with much less parameters, memory footprint and execution time.

\subsection{Pyramidal features extractor}

Input features are extracted by a small encoder architecture inspired by \cite{sun2017pwc}, made of 12 convolutional layers. At full resolution, the first layer produces the first level of the pyramid by applying convolutions with stride 2 followed by a second convolutional layer. Adopting this scheme at each level the resolution is decimated down to the lowest resolution (highest level of the pyramid) producing a total of 6 levels, from L1 to L6, corresponding respectively to image resolution from half to $\frac{1}{64}$ of the original input size. Each down-sampling module produces a larger number of extracted features, respectively 16, 32, 64, 96, 128, and 192, and each convolutional layers deploys $3 \times 3$ kernels and is followed by a leaky ReLU with $\alpha = 0.2$. Despite the small receptive field, this \emph{coarse-to-fine} strategy allows us to include global context at the higher levels (i.e., lower image resolution) of the pyramid as well as to refine details at the lower levels (i.e., higher image resolution) and at the same time to significantly reduce the amount of parameters and memory footprint.

\subsection{Depth decoders and upsampling}

At the highest level of the pyramid, extracted features are processed by a depth decoder made of 4 convolutional layers, producing respectively 96, 64, 32 and 8 feature maps. The output of this decoder is used for two purposes: i) to extract a depth map at the current resolution, by means of a sigmoid operator and ii) to pass the processed features at the next level in the pyramid, by means of a $2 \times 2$ deconvolution with stride 2 which increases by a factor 2 the spatial resolution.
The next level concatenates the features extracted from the input frame with those up-sampled and process them with a new decoder, repeating this procedure up to the highest resolution level. Each convolutional layer uses $3 \times 3$ kernels and is followed, as for deconvolutional layers, by leaky ReLU activations, except the last one which is followed by a Sigmoid activation to normalize the outputs.
With such design, at each scale PyD-Net learns to predict depth at full resolution. We will show in the experimental results how this design strategy, up-sampling depth maps from lower resolution decoders, allows to quickly infer depth maps with accuracy comparable to state-of-the-art. Indeed, it requires only a subset of decoders at test time reducing memory requirements and runtime thus making our proposal suited for CPU deployment.

\begin{table*}[t]
\center
\begin{tabular}{|c|c|cccc|ccc|c|}
\cline{5-8}
\multicolumn{4}{c}{} & \multicolumn{2}{|c|}{\cellcolor{blue!25}Lower is better}
 & \multicolumn{2}{c|}{\cellcolor{LightCyan}Higher is better} & \multicolumn{2}{c}{} \\
\hline
Method & Training dataset & \cellcolor{blue!25} Abs Rel & \cellcolor{blue!25} Sq Rel & \cellcolor{blue!25} RMSE & \cellcolor{blue!25} RMSE log & \cellcolor{LightCyan}$\delta<1.25$ & \cellcolor{LightCyan}$\delta<1.25^2$ & \cellcolor{LightCyan}$\delta<1.25^3$ & Params. \\
\hline
Eigen et al. \cite{eigen2014depth} & K & 0.203$^5$ & 1.548$^4$ & 6.307$^4$ & 0.282$^5$ & 0.702$^4$ & 0.890$^5$ & 0.958$^5$ & 54.2M \\
Liu et al. \cite{liu2016learning} & K & 0.201$^4$ & 1.584$^5$ & 6.471$^5$ & 0.273$^4$ & 0.680$^5$ & 0.898$^4$ & 0.967$^1$ & 40.0M \\
Zhou et al. \cite{zhou2017unsupervised} & K & 0.208$^6$ & 1.768$^6$ & 6.856$^6$ & 0.283$^6$ & 0.678$^6$ & 0.885$^6$ & 0.957$^6$ & 34.2M \\
Godard et al. \cite{godard2017unsupervised} & K & 0.148$^1$ & 1.344$^1$ & 5.927$^1$ & 0.247$^1$ & 0.803$^1$ & 0.922$^1$ & 0.964$^2$ & 31.6M \\
PyD-Net (50) & K & 0.163$^3$ & 1.399$^3$ & 6.253$^3$ & 0.262$^3$ & 0.759$^3$ & 0.911$^3$ & 0.961$^4$ & 1.9M \\
PyD-Net (200) & K & 0.153$^2$ & 1.363$^2$ & 6.030$^2$ & 0.252$^2$ & 0.789$^2$ & 0.918$^2$ & 0.963$^3$ & 1.9M \\
\hline
Garg et al. \cite{garg2016unsupervised} cap 50m & K & 0.169$^4$ & 1.080$^4$ & 5.104$^4$ & 0.273$^4$ & 0.740$^4$ & 0.904$^4$ & 0.962$^4$ & 16.8M \\
\cline{10-10}
Godard et al. \cite{godard2017unsupervised} cap 50m & K & 0.140$^1$ & 0.976$^1$ & 4.471$^1$ & 0.232$^1$ & 0.818$^1$ & 0.931$^2$ & 0.969$^2$ \\
PyD-Net (50) cap 50m & K &  0.155$^3$ & 1.045$^3$ & 4.776$^3$ & 0.247$^3$ & 0.774$^3$ & 0.921$^3$ & 0.967$^3$ \\
PyD-Net (200) cap 50m& K & 0.145$^2$ & 1.014$^2$ & 4.608$^2$ & 0.227$^2$ & 0.813$^2$ & 0.934$^1$ & 0.972$^1$ \\
\cline{1-9}
Zhou et al. \cite{zhou2017unsupervised} & CS+K & 0.198$^4$ & 1.836$^4$ & 6.565$^4$ & 0.275$^4$ & 0.718$^4$ & 0.901$^4$ & 0.960$^4$ \\
Godard et al. \cite{godard2017unsupervised} & CS+K & 0.124$^1$ & 1.076$^1$ & 5.311$^1$ & 0.219$^1$ & 0.847$^1$ & 0.942$^1$ & 0.973$^1$ \\
PyD-Net (50) & CS+K & 0.148$^3$ & 1.316$^3$ & 5.929$^3$ & 0.244$^2$ & 0.800$^3$ & 0.925$^3$ & 0.967$^2$ \\
PyD-Net (200) & CS+K & 0.146$^2$ & 1.291$^2$ & 5.907$^2$ & 0.245$^3$ & 0.801$^2$ & 0.926$^2$ & 0.967$^2$ \\
\cline{1-9}
\end{tabular}
\caption{Evaluation on KITTI \cite{geiger2012we} using the split of Eigen et al. \cite{eigen2014depth}. For training, K refers to KITTI dataset, CS+K means training on CityScapes \cite{cordts2016cityscapes} followed by fine-tuning on KITTI as outlined in \cite{godard2017unsupervised}. On top and middle of the table evaluation of all existing methods trained on K, at the bottom evaluation of unsupervised methods trained on CS+K. We report results for PyD-Net with two training configurations.}
\label{tab:eigen}
\end{table*}

\subsection{Training loss}

We train PyD-Net to estimate depth at each resolution deploying a multi-scale loss function as sum of different contributions computed at scales $s \in [1..6]$

\begin{equation}
\mathcal{L}_s = \alpha_{ap}(\mathcal{L}^l_{ap} + \mathcal{L}^r_{ap}) + \alpha_{ds}(\mathcal{L}^l_{ds}+\mathcal{L}^r_{ds}) + \alpha_{lr}(\mathcal{L}^l_{lr}+\mathcal{L}^r_{lr})
\label{eq:loss}
\end{equation}

The loss signal computed at each level of the pyramid is a weighted sum of three contributions computed on left and right images and predictions as in \cite{godard2017unsupervised}. 
The first term represents the reconstruction error $\mathcal{L}_{ap}$, measuring the difference between the original image $I^l$ and the warped one $\tilde{I}^l$ by means of SSIM \cite{wang2004image} and L1 difference.

\begin{equation}
\mathcal{L}^l_{ap} = \frac{1}{N} \sum_{i,j} \alpha \frac{1 - SSIM(I^l_{i,j},\tilde{I}^l_{i,j})}{2} + (1-\alpha)||(I^l_{i,j},\tilde{I}^l_{i,j})||
\end{equation}

The disparity smoothness term, $\mathcal{L}_{ds}$, discourages depth discontinuities according to L1 penalty unless a gradient $\delta I$ occurs on the image.

\begin{equation}
\mathcal{L}^l_{ds} = \frac{1}{N} \sum_{i,j} |\delta_x d^l_{i,j}|e^{-||\delta_x I^l_{i,j}||} + |\delta_y d_{i,j}|e^{-||\delta_y I^l_{ij}||}
\end{equation}

The third and last term includes the left-right consistency check, a well-known cue from traditional stereo algorithms \cite{scharstein2002taxonomy}, enforcing coherence between predicted left $d^l$ and right $d^r$ depth maps.

\begin{equation}
\mathcal{L}^l_{lr} = \frac{1}{N} \sum_{i,j} |d^l_{i,j} - d^r_{i,j + d^l_{i,j}}|
\end{equation}

The three terms are also computed for right image predictions, as shown in Equation \ref{eq:loss}. As in \cite{godard2017unsupervised}, the right input image and predicted output are used only at training time, while at testing time our framework works as a monocular depth estimator.

\section{Implementation details and training protocol}

We implemented PyD-Net in TensorFlow \cite{abadi2016tensorflow} and for experiments we deployed a pyramid with 6 levels (i.e., from 1 to 6) producing depth maps at a maximum resolution of half the original input size, up-sampled at full resolution by means of bilinear interpolation. We adopt this strategy since in our experiments deploying levels up to full resolution with PyD-Net does not improve the accuracy significantly and increases the complexity of the network. With this setting, PyD-Net counts 1.9 million parameters and runs in about 15ms on a Titan X Maxwell GPU while \cite{godard2017unsupervised}, with the VGG model, counts 31 million parameters and requires 35ms. More importantly, our simpler model enables its effective deployment even on low-end CPUs aimed at embedded systems or smartphones. Source code is available at \url{https://github.com/mattpoggi/pydnet}. 

We assess the effectiveness of our proposal with respect to the result reported in \cite{godard2017unsupervised}. For a fair comparison with \cite{godard2017unsupervised} we train our network with the same protocol for 50 epochs on batches of 8 images resized to $512 \times 256$, using 30 thousand images from KITTI raw data \cite{geiger2012we}. Moreover, we also provide results training PyD-Net for 200 epochs, showing how the final accuracy increases. It is worth noting that a longer schedule does not improve the performance of \cite{godard2017unsupervised}, already reaching top performance after 50 epochs.  On a Titan X GPU training takes about, respectively, 10 and 40 hours. Note that \cite{godard2017unsupervised} requires 20 hours for 50 epochs. The weights for our loss terms are always set to $\alpha_{ap} = 1$ and $\alpha_{lr} = 1$, while left-right consistency weight is set to $\alpha_{ds} = 0.1 / r$, being $r$ the down-sampling factor at each resolution layer as suggested in \cite{godard2017unsupervised}.
The inferred maps are multiplied by $0.3 \times $ image width, producing an inverse depth map proportional to maximum disparity between the training pairs. We use Adam optimizer \cite{kinga2015method} with $\beta_1 = 0.9$, $\beta_2 = 0.999$, and $\varepsilon = 10^{-8}$. We used a learning rate of $10^{-4}$ for the first 60\% epochs, halved every 20\% epochs until the end.
We perform data augmentation by randomly flipping input images horizontally and applying the following transformations: random gamma correction in [0.8,1.2], additive brightness in [0.5,2.0], and color shifts in [0.8,1.2] for each channel separately.

In \cite{godard2017unsupervised} an additional post-processing step was proposed to filter out and replace artifacts near depth discontinuities and image borders induced by training on stereo pairs. However, it requires to forward the input image twice thus doubling the processing time and memory, for this reason we do not include it in our evaluation.


\section{Experimental results}

We evaluate PyD-Net with respect to state-of-the-art on the KITTI dataset \cite{geiger2012we}. In particular, we first test the accuracy of our model on a portion of the full KITTI dataset commonly used in this field \cite{eigen2014depth}, then we focus on performance analysis of PyD-Net on different hardware devices, highlighting how our model can run even on low-powered CPU at about 2 Hz still enabling satisfying results, even more accurate than most techniques known in literature.

\subsection{Accuracy evaluation on Eigen split}

We compare the performance of PyD-Net with respect to known techniques for monocular depth estimation using the same protocol of \cite{godard2017unsupervised}. To do so, we use a test split of 697 images as proposed in \cite{eigen2014depth}, covering a total of 29 scenes out of the 61 available from KITTI raw data. The remaining 32 scenes are used to extract 22600 frames for training as in \cite{garg2016unsupervised,godard2017unsupervised}. Velodyne 3D points are reprojected on the left input image to obtain ground-truth labels on which evaluate depth estimation. 
As in \cite{godard2017unsupervised}, all methods use the same crop as \cite{eigen2014depth} to be directly comparable. Table \ref{tab:eigen} reports extensive comparison with both supervised \cite{eigen2014depth,liu2016learning} and unsupervised methods \cite{zhou2017unsupervised,godard2017unsupervised}. 
To compare the performance of the considered methods we use metrics commonly adopted in this field \cite{eigen2014depth} and we split our experiments into four main comparisons that we are going to discuss in detail.

In the first part of Table \ref{tab:eigen}, we compare PyD-Net trained on 50 and 200 epochs to supervised works by Eigen et al. \cite{eigen2014depth} and Liu et al. \cite{liu2016learning}, as well as with other unsupervised techniques by Zhou et al. \cite{zhou2017unsupervised} and Godard et al. \cite{godard2017unsupervised}. We report for each method the amount of parameters and, for each metric, the rank with respect to all the considered models. 
Excluding \cite{godard2017unsupervised} we can notice how PyD-Net, with a very low number of parameters and with both training configurations, significantly outperforms all considered methods on all metrics with the exception of $\delta<0.125^3$ on which Liu et al. \cite{liu2016learning} is even better than \cite{godard2017unsupervised}. Compared to \cite{godard2017unsupervised}, our network is less accurate but training PyD-Net for 200 epochs yields almost equivalent results.

To compare with the results reported by Garg et al. \cite{garg2016unsupervised}, in the middle part of Table \ref{tab:eigen} we evaluate predicted maps up to a maximum depth of 50 meters as in \cite{godard2017unsupervised}. Despite the smaller amount of parameters, reduced by a factor 8+, our network outperforms \cite{garg2016unsupervised} with both training configurations and has performance very close, and even better with metrics $\delta<0.125^2$ and $\delta<0.125^3$ training for 200 epochs, to Godard et al. \cite{godard2017unsupervised} a network counting more than $16\times$ parameters. As for previous experiment we can notice that training PyD-Net for 200 epochs always yields better accuracy. 

In the third part of Table \ref{tab:eigen}, we compare the performance of PyD-Net with respect to Zhou et al. \cite{zhou2017unsupervised} and Godard et al. \cite{godard2017unsupervised} unsupervised frameworks when trained on additional data. In particular, we first train the network for 50 epochs on CityScapes dataset and then we perform a fine-tuning on KITTI raw data according to the learning rate schedule described before. We can notice how training on additional data is beneficial for all the networks substantially confirming the previous trend.
Godard et al. method outperforms all other approaches while training PyD-Net for 200 epochs yields overall best performance for this method. However, even training PyD-Net for only 50 epochs always enables to achieve a better accuracy compared to the much complex network by Zhou et al. \cite{zhou2017unsupervised}. 

To summarize, our lightweight PyD-Net architecture outperforms more complex state-of-the-art methods \cite{eigen2014depth,liu2016learning,zhou2017unsupervised,garg2016unsupervised} and has results in most cases comparable to top-performing approach \cite{godard2017unsupervised}. Therefore, in the next section we evaluate in detail the impact of our design with respect to this latter method in terms of accuracy  and execution time, on three heterogeneous hardware architectures, with different setting of the two networks.  


\begin{table}[t]
\center
\begin{tabular}{|c|c|ccc|}
\hline
& Power  & 250+ [W] & 91+ [W] & 3.5 [W] \\
\hline
Model & Res. & Titan X & i7-6700K & Raspberry Pi 3 \\
\hline
Godard et al. \cite{godard2017unsupervised} & F & 0.035 s & 0.67 s & 10.21 s\\
\hline
Godard et al. \cite{godard2017unsupervised} & H & 0.030 s & 0.59 s & 8.14 s\\
PyD-Net & H & 0.020 s & 0.12 s & 1.72 s \\
\hline
Godard et al. \cite{godard2017unsupervised} & Q & 0.028 s & 0.54 s & 6.72 s\\
PyD-Net & Q & 0.011 s & 0.05 s & 0.82 s\\
\hline
Godard et al. \cite{godard2017unsupervised} & E & 0.027 s & 0.47 s & 5.23 s\\
PyD-Net & E & 0.008 s & 0.03 s & 0.45 s \\
\hline
\end{tabular}
\caption{Runtime analysis. We report for PyD-Net and \cite{godard2017unsupervised} the average runtime required to process the same KITTI image with 3 heterogeneous architectures at Full, Half, Quarter and Eight resolution. The measured power consumption for the Raspberry Pi 3 concerns the whole system plus a Logitech HD C310 USB camera while for CPU and GPU it concerns only such devices.}
\label{tab:time}
\end{table}

\subsection{Runtime analysis on different architectures}

\begin{table*}[t]
\center
\begin{tabular}{|c|c|cccc|ccc|}
\cline{5-8}
\multicolumn{4}{c}{} & \multicolumn{2}{|c|}{\cellcolor{blue!25}Lower is better}
 & \multicolumn{2}{c|}{\cellcolor{LightCyan}Higher is better} & \multicolumn{1}{c}{} \\
\hline
Method & Res. & \cellcolor{blue!25}Abs Rel & \cellcolor{blue!25}Sq Rel & \cellcolor{blue!25}RMSE & \cellcolor{blue!25}RMSE log & \cellcolor{LightCyan}$\delta<0.125$ & \cellcolor{LightCyan}$\delta<0.125^2$ & \cellcolor{LightCyan}$\delta<0.125^3$\\
\hline
Godard et al. \cite{godard2017unsupervised} & F & 0.124 & 1.076 & 5.311 & 0.219 & 0.847 & 0.942 & 0.973 \\
\hline
Godard et al. \cite{godard2017unsupervised} & H & 0.126 & 1.051 & 5.347 & 0.222 & 0.843 & 0.940 & 0.972\\
PyD-Net (50) & H & 0.148 & 1.316 & 5.929 & 0.244 & 0.800 & 0.925 & 0.967\\
PyD-Net (200) & H & 0.146 & 1.291 & 5.907 & 0.245 & 0.801 & 0.926 & 0.967\\
\hline
Godard et al. \cite{godard2017unsupervised} & Q & 0.132 & 1.091 & 5.632 & 0.231 & 0.830 & 0.935 & 0.970\\
PyD-Net (50) & Q & 0.152 & 1.342 & 6.185 & 0.252 & 0.789 & 0.920 & 0.964\\
PyD-Net (200) & Q & 0.148 & 1.285 & 6.146 & 0.252 & 0.787 & 0.919 & 0.965\\
\hline
Godard et al. \cite{godard2017unsupervised} & E & 0.160 & 1.601 & 7.121 & 0.270 & 0.773 & 0.909 & 0.958\\
PyD-Net (50) & E & 0.169 & 1.659 & 7.161 & 0.280 & 0.751 & 0.901 & 0.954\\
PyD-Net (200) & E & 0.167 & 1.643 & 7.222 & 0.282 & 0.747 & 0.898 & 0.953\\
\hline
\end{tabular} \\
\caption{Comparison between \cite{godard2017unsupervised} and PyD-Net at different resolutions. All models were trained on CS+K datasets and results are not post-processed to achieve maximum speed. As for Table \ref{tab:eigen}, we report results for PyD-Net with two training configurations.}
\label{tab:variants}
\end{table*}

Having assessed the accuracy of PyD-Net with respect to state-of-the-art, we compare on different hardware platforms the performance of our network with the top-performing one by Godard et al. \cite{godard2017unsupervised} . 
The reduced amount of parameters makes our model much less memory demanding and much faster thus allowing for real-time processing even on CPUs. This fact is highly desirable since GPU acceleration is not always feasible in applications scenarios characterized by low power constraints. Moreover, the pyramidal structure depicted in Figure \ref{fig:architecture} infers depth at different levels, getting more accurate at the higher resolution. This also happens for other models producing multi-scale outputs \cite{godard2017unsupervised,zhou2017unsupervised}. Thus, given a trained instance of any of these models, we can process outputs up to a lower resolution (e.g., half or quarter) to reduce the amount of computations, memory requirements and runtime. Therefore, we'll also investigate the impact of such strategy in terms of accuracy and execution time for our method and \cite{godard2017unsupervised}.

Table \ref{tab:time} reports running time analysis for PyD-Net and Godard et al. \cite{godard2017unsupervised} models estimating depth maps at different resolutions and with different devices. More precisely, the target systems are a Titan X Maxwell GPU, an i7-6700K CPU with 4 cores (4.2 Ghz) and a Raspberry Pi 3 board (ARM v8 processor Cortex-A53 1.2 Ghz). We report single forward time at full (F), half (H), quarter (Q) and eight (E) resolution. Full image resolution is set to $256\times512$ as in \cite{godard2017unsupervised}. For PyD-Net we report results up to half resolution for the reason previously outlined. Moreover, results do not include the post-processing step proposed in \cite{godard2017unsupervised} since it would duplicate the execution time and memory with small improvements in terms of accuracy.
First of all, we can notice how the model by Godard et al. \cite{godard2017unsupervised} is very fast on the high-end Titan X GPU while its performance drops dramatically when running on the Intel i7 CPU falling below 2 Hz. Moreover, it becomes unsuited for practical deployment on embedded CPUs such as the ARM processor of the Raspberry Pi 3. In this latter case it requires more than 10 seconds to process a single depth map at full resolution.
We can also notice how early stopping of the network to infer depth at reduced resolution leads to a very small decrease of running time for this method hardly bringing the framerate above 2 Hz on the Intel i7 and requiring more than 5 seconds on a Raspberry Pi 3 even stopping at $\frac{1}{8}$ resolution.
Looking at PyD-Net, even at the highest resolution H it takes 120 ms on the Intel i7 and less than 2 s on the ARM processor leading to $5\times$ speed up with respect to \cite{godard2017unsupervised} at the same resolution.
Moving to lower resolutions PyD-Net runs at 20 and 40 Hz, respectively, at Q and E resolutions yielding a speed up of $11\times$ and $18\times$ with respect to \cite{godard2017unsupervised}. Moreover, PyD-Net breaks the 1 Hz barrier even on the Raspberry Pi 3, with 1.2 and 2.2 Hz and a speed up w.r.t. \cite{godard2017unsupervised} of $8\times$ and $11\times$, respectively, at Q and E resolutions. On the same platform, equipped with 1 GB of RAM, our model requires 200, 150 and 120 MB, respectively, at H, Q and E resolution while the Godard et al. model about 275 MB at any resolution thus leaving a significantly smaller amount of memory available for other purposes. 

\begin{figure}[t]
\centering
 \setlength\tabcolsep{1.5pt}
\begin{tabular}{cccc}
\includegraphics[width=1.9cm]{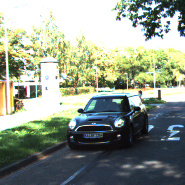} &
\includegraphics[width=1.9cm]{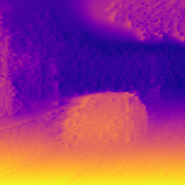} &
\includegraphics[width=1.9cm]{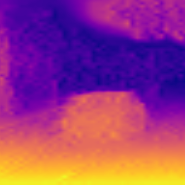} &
\includegraphics[width=1.9cm]{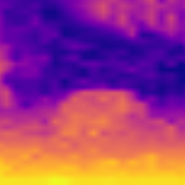} \\
\includegraphics[width=1.9cm]{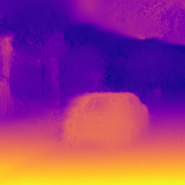} &
\includegraphics[width=1.9cm]{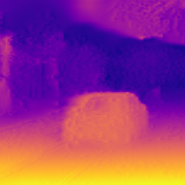} &
\includegraphics[width=1.9cm]{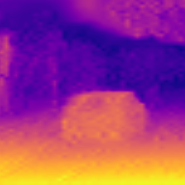} &
\includegraphics[width=1.9cm]{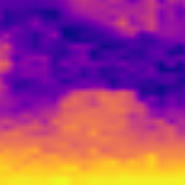} \\
a) & b) & c) & d) \\
\end{tabular}
\caption{Qualitative comparison on a portion of a KITTI image between PyD-net (top) and Godard et al. \cite{godard2017unsupervised} (bottom) respectively at F, H, Q and E resolution. Detailed timing analysis at each scale is reported in Table \ref{tab:time}.}
\label{fig:resolutions}
\end{figure}

These experiments highlight how PyD-Net enables, at the cost of small loss in accuracy, real-time performance on a standard CPU and it is also suited for practical deployment on devices with embedded CPUs. To better assess the trade-off between accuracy and execution time we report in Table \ref{tab:variants} detailed experimental results concerning PyD-Net and \cite{godard2017unsupervised} with different configurations/resolution. Results in the table were obtained from models trained on CS+K and evaluated on Eigen split \cite{eigen2014depth}.
We can observe how at E resolution PyD-Net performs similarly to the model proposed by Godard et al. \cite{godard2017unsupervised} providing output of the same dimensions. However, the gain in terms of runtime is quite high for PyD-Net as highlighted in the previous evaluation. In particular our competitor barely breaks the 1 Hz barrier on the i7 CPU and it is far behind on the Raspberry Pi, while PyD-Net runs, respectively, at 40 fps and about 2 fps on the same platforms.
As expected, from Table \ref{tab:variants}, stopping at lower resolution we can observe a loss in accuracy for both methods. However, it is worth to note that such reduction is more gradual for our network. Moreover, at E resolution the accuracy of Godard et al. network is substantially equivalent to PyD-Net with the advantages in terms of execution time previously discussed and reported in Table \ref{tab:time}. Finally, from the table we can also notice that even at the lowest resolution E, PyD-Net outperforms all remaining methods \cite{eigen2014depth,liu2016learning,zhou2017unsupervised,garg2016unsupervised} working at full resolution reported in Table \ref{tab:eigen}. 
Figure \ref{fig:resolutions} reports a qualitative comparison between PyD-Net and Godard et al. \cite{godard2017unsupervised} outputs at different resolutions.

The detailed evaluation reported proves that the proposed method can be effectively deployed on CPUs and actually it represents, to the best of our knowledge, the first architecture suited for CPU-based embedded systems enabling, for instance, its effective deployment with a Raspberry Pi 3 and a USB camera using a standard power bank for smartphones.
Moreover, despite its reduced complexity it enables unsupervised training and outperforms almost all methodologies proposed in literature for monocular depth estimation including supervised ones.

\section{Conclusions and future work}

In this paper we proposed PyD-Net, a novel and efficient architecture for unsupervised monocular depth estimation. As state-of-the-art method \cite{godard2017unsupervised}, it can be trained in unsupervised manner on rectified stereo pairs enabling comparable accuracy. However, the peculiar design of our network makes it suited for real-time applications on standard CPUs and also enables its effective deployment on embedded systems. Moreover, simplified configurations of our network allow to infer depth map at about 2 Hz on a Raspberry Pi 3 with accuracy higher than most state-of-the-art methods.
Future work is aimed at mapping PyD-Net on embedded devices specifically tailored for computer vision applications, such as the Intel Movidius NCS, thus paving the way for real-time monocular depth estimation in applications with hard low-power constraints (e.g., UAVs, wearable and assistive systems, etc).

\section*{Acknowledgment}

We gratefully acknowledge the support of NVIDIA Corporation with the donation of the Titan X GPU used for this research. We also thank Andrea Guccini for Figure \ref{fig:architecture}.

\bibliographystyle{IEEEtran}
\bibliography{egbib}

\end{document}